\documentclass[10pt,twocolumn,letterpaper]{article}

\usepackage[pagenumbers]{iccv}

\definecolor{iccvblue}{rgb}{0.21,0.49,0.74}
\usepackage[pagebackref,breaklinks,colorlinks,allcolors=iccvblue]{hyperref}
\usepackage{bm}
\usepackage{ulem}
\usepackage{multirow}
\usepackage{graphicx}
\usepackage{colortbl}
\usepackage{algorithm}
\usepackage{algorithmic}
\usepackage{svg}
\usepackage{CJKutf8}

\definecolor{blue}{RGB}{48,85,151}

\newcommand{\conf}[1]{\scalebox{0.7}{\textcolor{gray}{({#1})}}}

\def\paperID{***}
\def\confName{ICCV}
\def\confYear{2025}

\title{AdaViP: Aligning Multi-modal LLMs via Adaptive \\ Vision-enhanced Preference Optimization}
\author{
Jinda Lu \footnote[1]\ \ , 
Jinghan Li \footnote[1]\ \ , 
Yuan Gao \footnote[2]\ \ , 
Junkang Wu, 
Jiancan Wu, 
Xiang Wang \footnote[2]\ \ , 
Xiangnan He\\
\small 
University of Science and Technology of China  \\
{\tt\small \{lujd, lijh111, yuanga, wujk\}@mail.ustc.edu.cn, }\\ 
{\tt\small \{wujcan, xiangwang1223, xiangnanhe\}@gmail.com}\\
}
\begin{document}
\maketitle
\begin{abstract}
Preference alignment through Direct Preference Optimization (DPO) has demonstrated significant effectiveness in aligning multimodal large language models (MLLMs) with human preferences. 
However, existing methods focus primarily on language preferences while neglecting the critical visual context. 
In this paper, we propose an \textbf{Ada}ptive \textbf{Vi}sion-enhanced \textbf{P}reference optimization (AdaViP) that addresses these limitations through two key innovations: (1) vision-based preference pair construction, which integrates multiple visual foundation models to strategically remove key visual elements from the image, enhancing MLLMs' sensitivity to visual details; and (2) adaptive preference optimization that dynamically balances vision- and language-based preferences for more accurate alignment.
Extensive evaluations across different benchmarks demonstrate our effectiveness. Notably, our AdaViP-7B achieves 93.7\% and 96.4\% reductions in response-level and mentioned-level hallucination respectively on the Object HalBench, significantly outperforming current state-of-the-art methods. 
\end{abstract}    

\renewcommand{\thefootnote}{\fnsymbol{footnote}}
\footnotetext[1]{Equal Contributions.}
\footnotetext[2]{Corresponding authors.}

\section{Introduction}
\label{sec:intro}

\begin{figure}[t]
\begin{center}
\centerline{\includegraphics[width=\columnwidth]{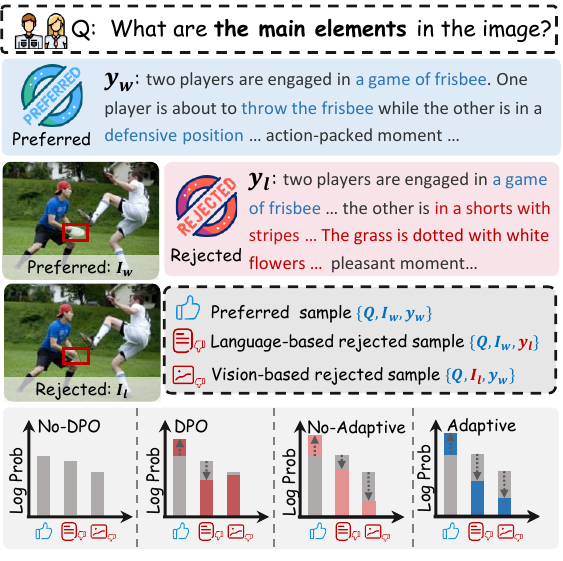}}
\caption{Comparison of preference construction methods and optimization strategies. (Top) The preferred and rejected images and responses. Language-based preferences maintain the same image ($I_w$) while varying responses ($y_w$ and $y_l$); Vision-based preferences keep the same response ($y_w$) while varying images ($I_w$ and $I_l$). (Bottom) Model log probabilities for different instances under different optimization approaches: base model (No-DPO), language-based DPO, fixed-weight vision-language DPO (No-Adaptive), and our adaptive strategy.}
\label{fig:figure-1}
\end{center}
\vskip -0.4in
\end{figure}

Multi-modal Large Language Models (MLLMs), building upon the foundation of Large Language Models (LLMs)~\cite{LLaVA, InternVL, QwenVL}, have demonstrated remarkable capabilities in visual understanding and achieved significant performance gains across various benchmarks.
However, hallucination --- where generated responses contradict visual references --- remains a significant challenge for broader MLLM applications~\cite{DAMA, VCD}. To address this limitation, Direct Preference Optimization (DPO)~\cite{DPO} has been adapted to multi-modal scenarios, emerging as an effective approach for aligning MLLM outputs with human preferences and demonstrating promising performances~\cite{RLAIF-V, mDPO}.

Current DPO methods for MLLMs~\cite{RLHF-V, RLAIF-V} focus primarily on constructing language-based preferences. These approaches typically utilize preference pairs $\{Q, I, y_w\}$ and $\{Q, I, y_l\}$, where $Q$ denotes the input query, $I$ represents the image, $y_w$ is the preferred response that aligns better with the visual content, and $y_l$ is the less preferred response. However, recent studies~\cite{mDPO, V-dpo} indicate that this approach may overemphasize language preferences while inadequately addressing visual context (\textit{ cf.} the stationary log probability of vision-based instances in Figure~\ref{fig:figure-1} DPO). Consequently, it is crucial to enhance MLLMs' visual attention capabilities beyond language-based preferences, improving their sensitivity to key visual details, and ultimately alleviating the hallucination issue.

To address this challenge, we propose an \textbf{Ada}ptive \textbf{Vi}sion-enhanced \textbf{P}reference optimization (AdaViP). Our method extends beyond traditional language-based preference pairs by introducing vision-based preference pairs $\{Q, I_w, y\}$ and $\{Q, I_l, y\}$, that precisely highlights the key visual elements. This framework enhances MLLMs' attention to key visual context, leading to outputs that are both more accurate and contextually relevant to the visual context.
Furthermore, our approach adaptively integrates visual and language-based preferences during optimization, enabling more effective cooperation between the two preference types and significantly improving alignment performance. Specifically, our AdaViP consists of:

\noindent \textbf{Vision-based preference pair construction (Sec.~\ref{subsec:vision_preference}).}
To construct vision-based preference pairs, we maintain the original preferred sample, while creating the rejected sample by removing key visual elements from the corresponding image.
To ensure accurate element removal, we leverage multiple visual foundation models~\cite{RAM, Grounding_DINO, SAM, LAMA}. Specifically, we employ the Recognize Anything Model \cite{RAM}, GroundingDINO \cite{Grounding_DINO} and Segment Anything Model (SAM) \cite{SAM} for instance recognition and segmentation, followed by visual inpainting~\cite{LAMA} to remove the identified objects. 
This process generates an instance-wise candidate set of perturbed images, from which we select the image with the least similarity to the preferred response $y_w$ as the rejected image.
As shown in Figure~\ref{fig:figure-1} (Top), the rejected image removes the ``frisbee'' instance based on the similarity to $y_w$. By contrasting this rejected sample with the preferred one in the preference pair, the MLLM learns to recognize and attend to subtle visual differences, thereby enhancing its sensitivity to key visual details.

\noindent \textbf{Adaptive Preference Optimization (Sec.~\ref{subsec:adaptive_preference}).}
As illustrated in Figure~\ref{fig:figure-1} (Bottom), the vision-based rejected sample has a lower log probability than the language-based one in ``No-DPO'', indicating that it is more easily identifiable as rejected. During ``No-Adaptive'' vision-language DPO training, MLLMs can quickly distinguish between preferred and vision-based rejected samples. 
This rapid differentiation results in larger gradients for the easily identifiable samples, while the more challenging samples receive less attention. 
Consequently, this leads to an imbalanced reduction in the log probabilities of vision- and language-based rejected samples, favoring the easily identifiable one. 
{To address this imbalance, we propose an adaptive strategy to dynamically modulate the effects of vision- and language-based preference pairs. By aligning these preferences into a unified distribution, our approach mitigates the disparities, and ensures a more contextually accurate and effective alignment.}
\noindent Our contributions can be summarized as:
\begin{itemize}
    \item We propose a vision-based preference pair construction, by collaborating with various visual foundation models, we precisely extract and remove the key visual elements, enhancing the MLLM's sensitivity to visual context.
    \item We introduce an adaptive preference optimization, by dynamically balancing the influences of vision- and language-based preferences, ensuring a more robust and accurate alignment.
    \item We conduct extensive empirical evaluations over three distinct benchmarks, involving both discriminative and generative tasks. Our approach demonstrates significant alignment performance improvements.
\end{itemize}
\section{Preliminaries}
\label{sec:preliminary}
In this section, we revisit the preference learning procedure of multi-modal large language models.

\noindent \textbf{Preference Data Construction.}
Preference learning of an MLLM $\bm{\pi}$ starts by sampling preference data from a Supervised Fine-Turned (SFT) model $\bm{\pi}_{\text{SFT}}$. 
Specifically, The SFT process enables the MLLM to align with multi-modal tasks by fine-tuning the MLLM with well-curated millions of multi-modal question-answer instances ($Q, \mathcal{I}, y$) $\sim$ $\mathcal{D}_{\text{SFT}}$, where ($Q, \mathcal{I}, y$) represents the question, image, and answer, respectively. 
Subsequently, the preference pair can be sampled from the SFT model as $(y_w, y_l) \sim \bm{\pi}_{\mathrm{SFT}}(y|Q,\mathcal{I})$, where $(y_w, y_l)$ are labeled as preferred and less preferred responses by either humans or AI models, formalized as $(y_w \succ  y_l | Q, \mathcal{I})$. The preference dataset $\mathcal{D}$ is constructed by repeating the sampling and labeling steps.

\begin{figure*}[t]
\begin{center}
\centerline{\includegraphics[width=\textwidth]{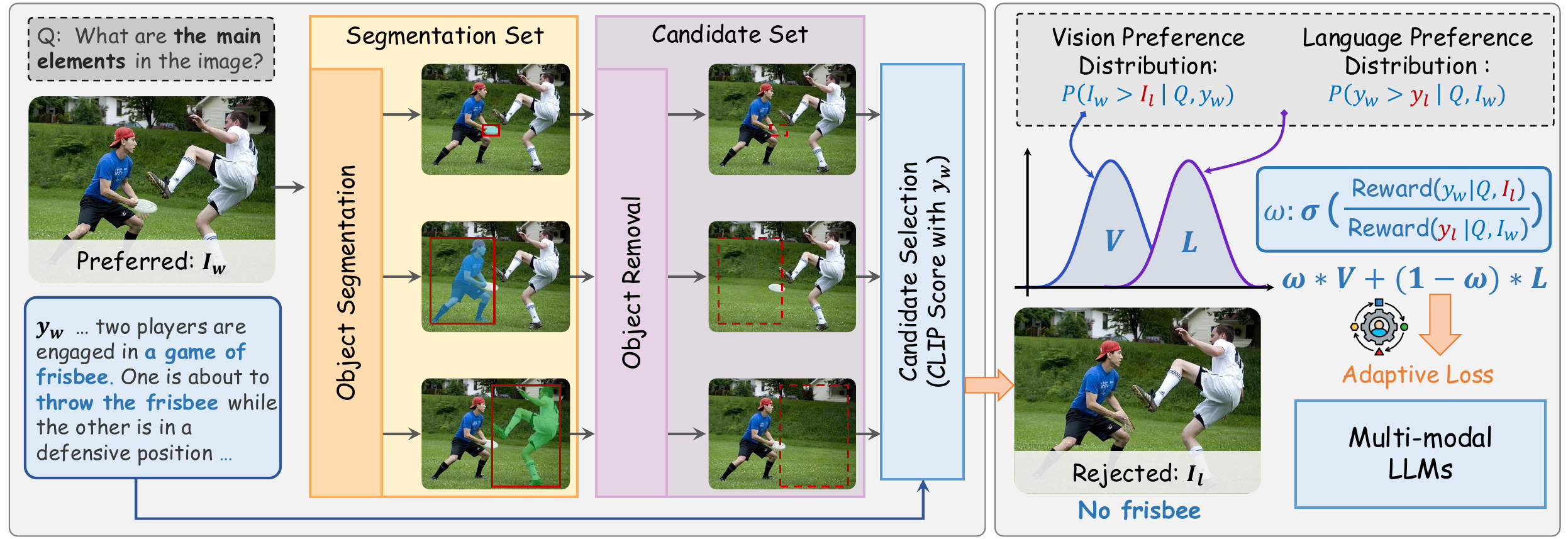}}
\caption{Overview of our \textbf{Ada}ptive \textbf{Vi}sion-aware \textbf{P}reference optimization (AdaViP). Given a preferred sample, the vision-based rejected one is constructed by locating and removing key elements of the image (Left). Subsequently, to effectively balance the vision- and language-based preference pair, we propose an adaptive loss that modulates the optimization procedure based on the relative rewards (right).}
\label{fig:pipleine}
\end{center}
\vskip -0.5cm
\end{figure*}

\noindent \textbf{Preference Learning with Reward Model.}
Given preference pair $(y_w, y_l) \sim \bm{\pi}_{\mathrm{SFT}}(y|Q,\mathcal{I})$, the preference learning process can be described in two stages: (1) reward modeling to learn the preference distribution and (2) preference optimization to optimize $\bm{\pi}_{\mathrm{SFT}}$ with the learned reward model.

To achieve reward modeling, pioneering work \cite{rlhf} employs the Bradley-Terry model \cite{BT_model} to model the preference distribution as:
\begin{equation}
\resizebox{.9\hsize}{!}{
\begin{math}
\begin{aligned}
\label{equ:probability}
    \mathrm{P}(y_w \succ  y_l|Q,\mathcal{I}) & =  \sigma(r^{*}(y_w|Q,\mathcal{I})- (r^{*}(y_l|Q,\mathcal{I})) \\
     & = \frac{\mathrm{exp}(r^{*}(y_w|Q,\mathcal{I}))}{\mathrm{exp}(r^{*}(y_w|Q,\mathcal{I}))+\mathrm{exp}(r^{*}(y_l|Q,\mathcal{I}))},
\end{aligned}
\end{math}
}
\end{equation}
where $r^{*}(y|Q,\mathcal{I})$ represents the expected reward model. And then, we minimize the $-\mathrm{logP}(y_w \succ y_l|Q, \mathcal{I})$ over the preference dataset $\mathcal{D}$ to learn the parametrized reward model $r_{\bm{\phi}}(y_w|Q,\mathcal{I})$. 
Subsequently, preference optimization can be achieved by employing policy optimization algorithms in RL such as PPO \cite{PPO} to maximize the learned reward model $r_{\phi}^{*}(y|Q,\mathcal{I})$, which can be formalized as:
\begin{equation}
\label{equ:ppo}
\begin{aligned}
    \underset{\bm{\pi}_{\theta}}{\text{max}} & \  \mathbf{E}_{(Q,\mathcal{I}) \sim \mathcal{D}, y \sim \bm{\pi}_{\theta}(\cdot|Q,\mathcal{I})} [r_{\phi}^{*}(y|Q,\mathcal{I})] \\
    & -\beta \mathbb{D}_{\mathbf{KL}}[\bm{\pi}_{\theta}(y|Q,\mathcal{I})||\bm{\pi}_{\text{ref}}(y|Q,\mathcal{I})], 
\end{aligned}
\end{equation}
where $\bm{\pi}_{\theta}$ and $\bm{\pi}_{\text{ref}}$ represents the policy and reference model, respectively.
Moreover, the learned policy $\bm{\pi}_{\theta}$ and the fixed reference $\bm{\pi}_{\text{ref}}$ are initially parameterized as $\bm{\pi}_{\text{SFT}}$.

\noindent \textbf{Direct Preference Optimization (DPO).}
To simplify the preference learning process, researchers reveal that the parametric reward model can be substituted with an implicit reward function, and then propose the DPO \cite{DPO} method. Specifically, the DPO loss can be described as:
\begin{equation}
\label{equ:dpo}
\begin{aligned}
    \mathcal{L}_{\text{dpo}} = - \bm{\mathrm{E}}_{(Q,\mathcal{I}, y_{w}, y_{l})} \bigg[ {\log \sigma}( & \beta \log \frac{{\pi}_{\bm{\theta}}(y_{w}|Q,\mathcal{I})}{{\pi}_{\text{ref}}(y_{w}|Q,\mathcal{I})} \\
    - & \beta \log \frac{{\pi}_{\bm{\theta}}(y_{l}|Q,\mathcal{I})}{{\pi}_{\text{ref}}(y_{l}|Q,\mathcal{I})}) \bigg].
\end{aligned}
\end{equation}
The implicit reward function $\hat{r}(y|Q,\mathcal{I})$ is formalized as: 
\begin{equation}
\label{equ:dpo_reward}
\begin{aligned}
    \hat{r}(y|Q,\mathcal{I}) =  \beta \log \frac{{\pi}_{\bm{\theta}}(y|Q,\mathcal{I})}{{\pi}_{\text{ref}}(y|Q, \mathcal{I})}.
\end{aligned}
\end{equation}
\section{Approach}
\label{sec:approach}
In this section, we elaborate on our AdaViP in detail. Specifically, we first briefly review several key visual foundation models, then illustrate the construction of vision-based preference pairs, and finally present our adaptive strategy for balancing vision- and language-based preferences. The algorithm of AdaViP is listed in Algorithm~\ref{algorithm}.

An overview of our AdaViP is presented in Figure \ref{fig:pipleine}. Given a preferred sample, we construct the corresponding vision-based rejected sample using visual foundation models from different aspects. We detect and remove key visual elements, then select the image with the least similarity to the preferred response $y_{w}$ as the rejected image.  Subsequently, we adaptively balance the learning process of vision- and language-based preferences by incorporating the relative reward values.

\subsection{Visual Foundation Models}
Leveraging large-scale data and advanced pre-training strategies, vision foundation models have achieved remarkable success, demonstrating significant open-set generalization and zero-shot transferability across diverse domains \cite{SAM, Grounding_DINO, LAMA, RAM}. By taking advantage of these models, we can precisely manipulate the image elements. In this section, we briefly review relevant models to provide a comprehensive context for subsequent discussions.

\noindent \textbf{Recognition.} Benefiting from web-scale image-text pairs, CLIP \cite{CLIP} $\Omega_{\text{CLIP}}$ has substantially enhanced the vision-language alignment capability. Building upon this, Recognize Anything \cite{RAM} $\Omega_{\text{RAM}}$ dramatically improves image tagging abilities, enabling precise automatic annotation of the image across comprehensive textual categories.

\noindent \textbf{Detection.} Inspired by larger backbones \cite{DETR, DINO} and extensive datasets \cite{Objects365}, Grounding DINO \cite{Grounding_DINO} $\Omega_{\text{Ground DINO}}$ advances open-set detection by integrating detection with grounding. Leveraging the CLIP text encoder \cite{CLIP}, it enables accurate object localization for specific textual categories when given an image with textual descriptions.

\noindent \textbf{Segmentation.} Employing billions of instances with masks and pre-training across diverse tasks, Segment Anything Model (SAM) \cite{SAM} $\Omega_{\text{SAM}}$ demonstrates remarkable generalization capabilities across distinct tasks with multiple input prompts. In this work, we employ the bounding boxes as the prompt for precise segmentation.

\noindent \textbf{Removal.} 
By utilizing larger-scale datasets with various scale inpainting masks, models such as Big LaMa \cite{LAMA} $\Omega_{\text{LaMa}}$ demonstrate exceptional generalization in object removal. In this work, leveraging the segmented masks, we meticulously preserve the underlying image structure and precisely remove the objects. 

\subsection{Construction of Vision-based Preference Pair}
\label{subsec:vision_preference}
In this section, we describe the construction of a vision-based preference pair. Given a language-based preference pair $(y_w \succ y_l | Q, \mathcal{I}) \sim \mathcal{D}$, we retain the preferred sample as $(y_w|Q, \mathcal{I}_{w})$ and generate the rejected sample by systematically perturbing the original image $\mathcal{I}_{w}$. Specifically, we denote the vision-based rejected sample as $(y_w | Q, \mathcal{I}_{l})$, which is constructed through the following steps.

We cascade the visual foundation models to construct the candidate set of $\mathcal{I}_{l}$. Specifically, we begin by extracting the textual category set $\mathcal{T}$ of the image $\mathcal{I}_{w}$ using $\Omega_{\text{RAM}}$:
\begin{equation}
\label{equ:RAM}
   \mathcal{T} = \Omega_{\text{RAM}}(\mathcal{I}_{w}), \,\,\mathcal{T} = \{ t_{i} \mid i = 1,\ldots,n \},
\end{equation}
where $n$ is the category number. 
Nextly, we generate the corresponding bounding box set via $\Omega_{\text{Ground-DINO}}$:
\begin{equation}
\label{equ:grounding_dino}
    \mathcal{B} = \{ b_{i} = \Omega_{\text{Ground-DINO}}(\mathcal{I}_{w}, t_{i}) \mid i = 1,\ldots,n \},
\end{equation}
where each box has 4 points, and $\mathcal{B} \sim \mathbb{R}^{n \times 4}$. With the bounding box set $\mathcal{B}$, we can easily obtain each segmentation mask $\mathcal{M}$ with $\Omega_{\text{SAM}}$:
\begin{equation}
\label{equ:SAM}
    \mathcal{M} = \{ m_{i} = \Omega_{\text{SAM}}(\mathcal{I}_{w}, b_{i}) \mid i = 1,\ldots,n \}.
\end{equation}
Then, with the segmented mask set $\mathcal{M}$, we precisely remove elements from the image, collecting the modified images in a candidate set $\mathcal{C}$ by $\Omega_{\text{LaMa}}$:
\begin{equation}
\label{equ:lama}
    \mathcal{C} = \{ c_{i} = \Omega_{\text{LaMa}}(\mathcal{I}_{w}, m_{i}) \mid i = 1,\ldots,n \}.
\end{equation}

To select the most representative candidate from $\mathcal{C}$, we measure the similarity between each candidate and the preferred response $y_{w}$ with the CLIP classifier $\Omega_{\text{CLIP}}$. We address CLIP's token length constraints by decomposing $y_{w}$ into self-contained sentences using LLaMA-3 \cite{LLaMA3}, denoted as $\{s_{w,j}\}_{j=1}^{m}$. 
For each candidate, we calculate the CLIP score by summing the scores across these sentences:
\begin{equation}
\label{equ:clip}
    \text{Score}_{i} = \sum_{j=1}^{n}{\Omega}_{\text{CLIP}}(c_{i}, s_{w,j}).
\end{equation}
The rejected image is selected as the candidate with the lowest CLIP score. 
Unlike existing methods \cite{V-dpo, POVID, mDPO}, we accurately perturb the image by removing key elements, maintaining the image structure, and mitigating randomness during the perturbation process, thus ensuring a more robust and consistent rejected image construction.

\subsection{Adaptive Preference Optimization}
\label{subsec:adaptive_preference}
In this section, we introduce the adaptive learning procedure that uses vision- and language-based preference pairs. Specifically, after constructing vision-based preference data, the preference dataset $\mathcal{D}$ comprises $\{ Q, \mathcal{I}_{w}, \mathcal{I}_{l}, y_w, y_l \}$. For clarity, we denote the preferred sample as $(Q, \mathcal{I}_{w}, y_w) \sim \mathcal{D}^{\scriptscriptstyle++}$, the vision-based rejected sample as $(Q, \mathcal{I}_{l}, y_w) \sim \mathcal{D}^{\scriptscriptstyle-+}$, and the language-based rejected sample as $(Q,\mathcal{I}_{w}, y_{l}) \sim \mathcal{D}^{\scriptscriptstyle+-}$. We use $\mathcal{D}^{\scriptscriptstyle++}$, $\mathcal{D}^{\scriptscriptstyle-+}$, and $\mathcal{D}^{\scriptscriptstyle+-}$ to represent the preferred, vision-based, and language-based rejected data, respectively.

Preference learning traditionally relies on pair-wise comparisons, existing methods \cite{V-dpo, mDPO} decompose multi-modal preferences into separate distributions. Specifically, they model the vision-based preference distribution as $\text{P}(\mathcal{D}^{\scriptscriptstyle++} \succ \mathcal{D}^{\scriptscriptstyle-+})$, and language-based preference distribution as $\text{P}(\mathcal{D}^{\scriptscriptstyle++} \succ \mathcal{D}^{\scriptscriptstyle+-})$. Similar to multi-task learning \cite{Gradient_surgery}, they jointly learn preferences using carefully designed hyperparameters or even additional regularization terms to ensure balance. 

However, such methods exhibit pronounced sensitivity to hyperparameters or regularization terms. To address this limitation, inspired by the Plackett-Luce model \cite{PL_Model, softmax_dpo}, we unify the vision- and language-based preferences as: 
\begin{equation}
    \text{P}(\mathcal{D}^{{\scriptscriptstyle++}} \succ \{\mathcal{D}^{\scriptscriptstyle+-}, \mathcal{D}^{\scriptscriptstyle-+}\}) = \frac{{\rm exp}(r^{*}(\mathcal{D}^{\scriptscriptstyle++}))}{\sum_{j \in \mathcal{J}}{\rm exp}(r^{*}(\mathcal{D}^{j}))},
\end{equation}
where $\mathcal{J}$ represents the preference indicators, and $\mathcal{J}=\{ {\scriptscriptstyle++}, {\scriptscriptstyle+-}, {\scriptscriptstyle-+}\}$. 
Subsequently, by substituting the expected reward model $r^{*}$ with the implicit reward $\hat{r}$ from Equation \ref{equ:dpo_reward}, we derive the final loss to adaptively optimize both the vision- and language-based preferences:
\begin{equation}
\begin{aligned}
\label{equ:loss}
    & \mathcal{L}(\pi_\theta;\pi_{\rm ref})  =-\mathbb{E}_{(\mathcal{D}^{\scriptscriptstyle++},\mathcal{D}^{\scriptscriptstyle+-},\mathcal{D}^{\scriptscriptstyle-+})} \bigg[{\rm log}\,\sigma 
    \bigg(-\rm{log}  \\  
    & \sum_{\substack{j \in \mathcal{J} \\ j \neq {\scriptscriptstyle++}}}{\rm exp} \big( \beta\,\rm{log} \,\frac{\pi_{\bm{\theta}}(\mathcal{D}^{j})}{\pi_{\rm{ref}}(\mathcal{D}^{j})} - \beta\,\rm{log} \,\frac{\pi_{\bm{\theta}}(\mathcal{D}^{\scriptscriptstyle++})}{\pi_{\rm{ref}}(\mathcal{D}^{\scriptscriptstyle++})}\big) \bigg) \bigg]. 
\end{aligned}
\end{equation}
Finally, we systematically analyze how such a simple strategy can enhance adaptive optimization across different preferences. Specifically, the gradient concerning the parameters $\theta$ can be written as:
\begin{equation}
\begin{aligned}
\label{equ:gradient}
    \nabla_\theta &\mathcal{L}(\pi_\theta;\pi_{\rm ref})  =  -\beta\mathbb{E}_{(\mathcal{D}^{\scriptscriptstyle++},\mathcal{D}^{\scriptscriptstyle+-},\mathcal{D}^{\scriptscriptstyle-+})}\bigg[\\
    &
    \underbrace{\sigma( \rm{log}\,\sum_{\substack{j \in \mathcal{J} \\ j \neq {\scriptscriptstyle++}}} \rm{exp}(\hat{r}(\mathcal{D}^{j})-\hat{r}(\mathcal{D}^{\scriptscriptstyle++}))}_{\text{higher weight if either reward estimation is wrong}} \\
    &
    \big[
    \underbrace{\nabla_\theta\,{\rm log}\pi_\theta(\mathcal{D}^{\scriptscriptstyle++})}_{\text{increase}}- \underbrace{\sum_{\substack{j \in \mathcal{J} \\ j \neq {\scriptscriptstyle++}}} \omega_{j} \cdot \nabla_\theta\,{\rm log}\pi_\theta(\mathcal{D}^{\scriptscriptstyle j})}_{\text{weighted decrease}}
    \big]
    \bigg].
\end{aligned}
\end{equation}
In line with DPO~\cite{DPO}, our strategy also meets two key criteria: (1) the MLLM pays more attention to pairs where the implicit reward model rates the rejected sample higher than the preferred; (2) the implicit reward disparity between the preferred and rejected samples gradually increases. Furthermore, our adaptive loss mechanism also assigns different weights to each rejected sample. The weight for each preference can be mathematically derived as:
\begin{eqnarray}
\label{equ: weight}
\begin{aligned}
& \omega_{\scriptscriptstyle +-} = \text{Sigmoid}\left(\hat{r}(\mathcal{D}^{\scriptscriptstyle +-}) - {\hat{r}(\mathcal{D}^{\scriptscriptstyle -+})}\right),
 \\ 
&\omega_{\scriptscriptstyle -+} = \text{Sigmoid}\left({\hat{r}(\mathcal{D}^{\scriptscriptstyle -+})} - {\hat{r}(\mathcal{D}^{\scriptscriptstyle +-})}\right).
\end{aligned}
\end{eqnarray}
In this way, it enables the model to adaptively optimize the preferences by dynamically weighting them according to their reward value ratios, with larger ratios indicating more attention to the corresponding rejected sample.

\begin{algorithm}[t]
   \caption{Algorithm of AdaViP.}
   \label{algorithm}
\begin{algorithmic}
   \STATE {\bfseries Input:} 
   preferred dataset $\mathcal{D}^{\scriptscriptstyle++}$, language-based rejected dataset $\mathcal{D}^{\scriptscriptstyle+-}$, SFT model $\pi_{\text{SFT}}$, vision foundation models $ \{ \Omega_{\text{RAM}}, \Omega_{\text{Grounding DINO}}, \Omega_{\text{SAM}}, \Omega_{\text{LaMa}}, \Omega_{\text{CLIP}} \} $.
   \STATE Initialize vision-based rejected dataset $\mathcal{D}^{\scriptscriptstyle-+} \gets \emptyset $;
   \FOR{$\{ Q, \mathcal{I}_{w}, y_{w}\}$ $\sim$ $\mathcal{D}^{\scriptscriptstyle++}$}
        \STATE $\mathcal{M} = \{ \Omega_{\text{RAM},\text{Ground-DINO},\text{SAM}}\}(\mathcal{I}_{w})$;
         \COMMENT{Equ.~\eqref{equ:RAM} $\to$ \eqref{equ:SAM}};
        \STATE Candidate Set $\mathcal{C} = \Omega_{\text{LaMa}}(\mathcal{I}_{w}, \mathcal{M}) \}$
        \COMMENT{Equ.~\eqref{equ:lama}};
        \STATE $\text{Score} = {\Omega}_{\text{CLIP}}(\mathcal{C}, y_{w})$; \COMMENT{Equ.~\eqref{equ:clip}};
        \STATE $\mathcal{I}_{l} \gets $ Image with the least score in $\mathcal{C}$.
        \STATE $\mathcal{D}^{\scriptscriptstyle-+} \gets \mathcal{D}^{\scriptscriptstyle-+} \, \cup \{ Q, \mathcal{I}_{l}, y_{w}\}$;
    \ENDFOR
   \STATE Initialize model $\pi_{\bm{\theta}}$ and reference model $\pi_{\text{ref}}$ $\gets$ $\pi_{\text{SFT}}$.
   \REPEAT
   \FOR{$\mathcal{B} = \{ (Q^{(i)}, \mathcal{I}_{w}^{(i)}, y_{w}^{(i)}), $ $\,(Q^{(i)}, \mathcal{I}_{l}^{(i)}, y_{w}^{(i)}), $ $\,(Q^{(i)}, $ $\mathcal{I}_{w}^{(i)}, $ $y_{l}^{i})\}_{i=1}^{N} \sim $ $\{ \mathcal{D}^{\scriptscriptstyle++},   \mathcal{D}^{\scriptscriptstyle-+}, \mathcal{D}^{\scriptscriptstyle+-} \}$}
        \STATE Compute $\mathcal{L}(\pi_\theta;\pi_{\rm ref})$ {w.r.t.} $\pi_{\bm{\theta}}, \, \pi_{\rm ref}$; \COMMENT{Equ.~\eqref{equ:loss}};
        \STATE Compute $ \nabla_\theta \mathcal{L}(\pi_\theta;\pi_{\rm ref})$ {w.r.t.} $\pi_{\bm{\theta}, \, \pi_{\rm ref}}$; \COMMENT{Equ.~\eqref{equ:gradient}};
        \STATE Update $\pi_{\bm{\theta}}$ as $\pi_{\bm{\theta}} \gets \pi_{\bm{\theta}} - \eta \nabla_\theta \mathcal{L}(\pi_\theta;\pi_{\rm ref})$; 
    \ENDFOR
   \UNTIL{The optimization is converged.}
   \STATE {\bfseries Output}: The constructed vision-based rejected data $\mathcal{D}^{\scriptscriptstyle-+}$, The optimized model $\pi_{\bm{\theta}}$.
\end{algorithmic}
\end{algorithm}

\section{Experiment}
\label{sec:experiment}

\begin{table*}[t]
\caption{Performance comparisons with state-of-the-art methods over three different benchmarks. 
We report non-hallucination rates for both responses (Non-Rsp.) and mentioned (Non-Men.) on Object HalBench \cite{object_hallucination_benchmark}. 
AMBER Score \cite{AMBER} is calculated by $(100-CHAIR+F1)/2$.
HalRefersers to the Hallucination Rate for MMHal Bench \cite{LLaVA-RLHF}.
The best results of all methods are indicated in \textbf{bold}, and the second best results are \uline{underlined}.
All compared results are sourced from \cite{RLAIF-V, TPO, mDPO}, and the reported results of LLaVA-1.5, DPO. 
Our AdaViP is evaluated using GPT-4-turbo-2024-04-09.}
\label{tab:comparison_SOTA}
\begin{center}
\scalebox{0.87}{
\begin{tabular}{l | c | c c | c c | c c | c | c c }
\toprule
\multirow{3}{*}{\textbf{Method}} & \multirow{3}{*}{\textbf{Size}} & \multicolumn{2}{c|}{\hspace{-1mm}\textbf{Object}} & \multicolumn{5}{c|}{{\hspace{-3mm}\textbf{AMBER}}} & \multicolumn{2}{c}{\textbf{MMHal-}}  \\

\cline{5-9}& & \multicolumn{2}{c|}{\hspace{-1mm}\textbf{HalBench}} & \multicolumn{2}{c|}{\hspace{-1mm}\textbf{Discriminative}} & \multicolumn{2}{c|}{\hspace{-1mm}\textbf{Generative}} & \multirow{2}{*}{\textbf{Score}($\uparrow$)} & \multicolumn{2}{c}{\textbf{Bench}} \\

\cline{3-8} \cline{10-11} & & Non-Rsp.($\uparrow$) & Non-Men.($\uparrow$) & Acc($\uparrow$) & F1($\uparrow$) & CHAIR($\downarrow$) & Cog.($\downarrow$) & & Scores($\uparrow$) & Hall.($\downarrow$) \\

\midrule
\multicolumn{8}{l}{\multirow{1}*{\textbf{Method (Hallucination-specific)}}}\\
\midrule
VCD~\conf{CVPR'24} & 7B & 51.2 & 75.7 & 71.8 & 74.9 & - & - & - & 2.12 & 54.2 \\
Less-is-more~\conf{ACL'24} & 7B & 59.7 & 82.2 & 72.4 & 75.8 & 5.1 & 2.0 & 85.4 & 2.33 & 50.0 \\
OPERA~\conf{CVPR'24} & 7B & 54.9 & 77.7 & 75.2 & 78.3 & - & - & -& 2.15 & 54.2 \\
CCA-LLaVA~\conf{NIPS'24} & 7B & 53.3 & 76.2 & 77.7 & 81.9 & - & - & -& 1.92 & 61.5 \\
\midrule
\multicolumn{8}{l}{\multirow{1}*{\textbf{Method (Preference optimization)}}}\\
\midrule
HA-DPO~\conf{arXiv'23} & 7B & 60.1 & 80.1 & 75.2 & 79.9 & 6.7 & 3.3 & 86.6 & 1.98 & 60.4 \\
POVID~\conf{arXiv'24} & 7B & 51.9 & 75.6 & \textbf{82.9} & \textbf{87.4} & 7.3 & 3.7 & 90.1 & 2.08 & 56.2 \\
RLAIF-V~\conf{arXiv'24} & 7B & 89.5 & 94.8 & 76.8 & 84.5 & 3.1 & \uline{1.0} & 90.7 & 2.95 & 32.3 \\
V-DPO~\conf{EMNLP'24} & 7B & - & - & - & 81.6 & 5.6 & 2.7 & 88.0 & 2.16 & 56.0 \\
MDPO~\conf{EMNLP'24} & 7B & 64.3 & 90.2 & - & - & 4.4 & 2.4 & - & 2.39 & 54.0 \\
LLaVA-RLHF~\conf{ACL'24} & 13B & 61.9 & 81.1 & 79.7 & 83.9 & 7.7 & 4.4 & 88.1 & 2.02 & 62.5 \\
RLHF-V~\conf{CVPR'24} & 13B & 87.8 & 92.5 & 72.6 & 75.0 & 6.3 & 2.1 & 84.4 & 2.45 & 51.0 \\
AMP-MEG~\conf{NIPS'24} & 13B & 68.3 & 79.4 & 79.5 & 84.6 & - & - & - & \uline{3.08} & 36.5 \\
\midrule
LLaVA-1.5 & 7B & 47.8 & 71.2 & 73.9 & 77.7 & 7.8 & 4.2 & 85.0 & 1.95 & 63.5 \\
\hspace{1mm} + DPO & 7B & 80.1 & 90.8 & 64.8 & 78.8 & 3.5 & \uline{1.0} & 87.7 & 2.00 & 42.0 \\
\hspace{1mm} + AdaViP & 7B & \uline{93.7} & \uline{96.4} & \uline{79.9} & \uline{85.8} & \textbf{1.8} & \textbf{0.3} & \textbf{92.0} & \textbf{3.12} & \textbf{28.0} \\
\midrule
LLaVA-1.5 & 13B & 50.0 & 76.4 & 71.2 & 73.0 & 7.8 & 4.2 & 82.6 & 2.36 & 56.0 \\
\hspace{1mm} + DPO & 13B & 82.3 & 91.3  & 66.5 & 78.7 & \uline{3.4} & 1.2 & 87.7 & 2.61 & \uline{30.0} \\
\hspace{1mm} + AdaViP & 13B & \textbf{95.4} & \textbf{97.7}  & 76.4 & 85.2 & \textbf{1.8} & \textbf{0.3} & \uline{91.7} & 2.85 & 34.0 \\
\midrule
\rowcolor{lightgray} \textbf{GPT-4V} & - & 86.4 & 92.7 & 83.4 & 87.4 & 4.6 & 2.6 & 91.4 & 3.49 & 28.1 \\
\bottomrule
\end{tabular}
}
\end{center}
\end{table*}

In this section, we validate the effectiveness of our \textbf{Ad}aptive \textbf{Vi}sion-enhanced \textbf{P}reference optimization (AdaViP) by answering the following \textbf{R}esearch \textbf{Q}uestions (RQs):
\begin{itemize}
    \item  What are the influences of different components (\textit{e.g.} the vision-based preference pair construction, and the adaptive optimization strategy)$?$
    \item How does AdaViP perform compared with state-of-the-art methods, including hallucination-specific, and preference optimization-based approaches$?$
    \item What is the respective effect of the vision- and language-based preferences during the training procedure$?$
\end{itemize}

\subsection{Experimental Settings}
We first outline the experimental settings, encompassing the dataset, baselines, metrics, and implementation specifics.

\noindent \textbf{Dataset.}
We employ the preference data of the LLaVA 1.5 model from \cite{RLAIF-V}, where the language-based preference is annotated by the open-source LLaVA-Next 34B model \cite{LLaVA-Next}. Specifically, the dataset comprises 22k preference data, with 13K images allocated for training. 
We construct the vision-based preference by removing the elements in the 13K images and then compare the similarity with the preferred response in each preference to construct the corresponding rejected image.

\noindent \textbf{Baselines.}
We conduct a comparative assessment against state-of-the-art baselines across several categories:
\newline
(1) {Baselines targeted at hallucination problems.} In this category, our primary comparisons include VCD \cite{VCD}, Less-is-more \cite{Less_Is_More}, OPERA \cite{OPERA}, and CCA-LLaVA \cite{CCA-LLaVA}.
\newline 
(2) {Preference Optimization-based baselines.} In this category, we mainly compare with HA-DPO \cite{HA-DPO}, POVID \cite{POVID}, RLAIF-V \cite{RLAIF-V}, V-DPO \cite{V-dpo}, MDPO \cite{mDPO}, LLaVA-RLHF \cite{LLaVA-RLHF}, RLHF-V \cite{RLHF-V}, and AMP-MEG \cite{AMP-MEG}.
\newline
(3) {Proprietary baseline.} We incorporate GPT-4V as a robust reference to assess the performance disparity between open-source and proprietary commercial models.

\noindent \textbf{Evaluation metrics.}
To evaluate the performance of our AdaViP and other baselines, we conduct experiments across three widely used metrics for MLLMs, focusing on hallucination problems and reflecting trustworthiness.
\newline
(1) {Object HalBench} \cite{object_hallucination_benchmark} is a widely recognized benchmark designed to evaluate object hallucination in detailed image descriptions, measuring the percentage of hallucinated responses and the proportion of hallucinated object mentions. Here, we report the \textbf{non}-hallucination rates.
\newline
(2) {AMBER} \cite{AMBER} is a comprehensive multi-dimensional hallucination benchmark comprising over 15k samples, designed to evaluate both generative and discriminative tasks related to object, attribute, and relation hallucination. We report the Accuracy and F1 metric of its discriminative component besides CHAIR scores and cognition of its generative component, respectively.
\newline
(3) {MMHal-Bench} \cite{LLaVA-RLHF} assesses response-level hallucination rate and informativeness. It utilizes GPT-4 to compare model outputs with object labels and human responses.

\noindent \textbf{Implementation details.}
We use LLaVA-1.5 7B and 13B as the backbone for our experiments and employ full parameter-tuning over the preference dataset with $5000$ steps. In the interest of reproducibility, we adhere to the hyperparameters specified in the official LLaVA GitHub repository \footnote{https://github.com/haotian-liu/LLaVA}. Specially, the batch sizes are configured to be $32$ and $16$ for 7B and 13B models, respectively. The penalty hyperparameter $\beta$ is set to $0.1$, consistent with the methodologies outlined in \cite{DPO, RLAIF-V}. All the experimental evaluations are performed utilizing four A100 80GB GPUs.

\subsection{Ablation Studies (RQ1)}
In this section, we evaluate different components and special designs of our AdaViP, such as the vision-based preference pair construction and adaptive optimization strategy. 

\begin{figure}[t]
\begin{center}
\centerline{\includegraphics[width=\columnwidth]{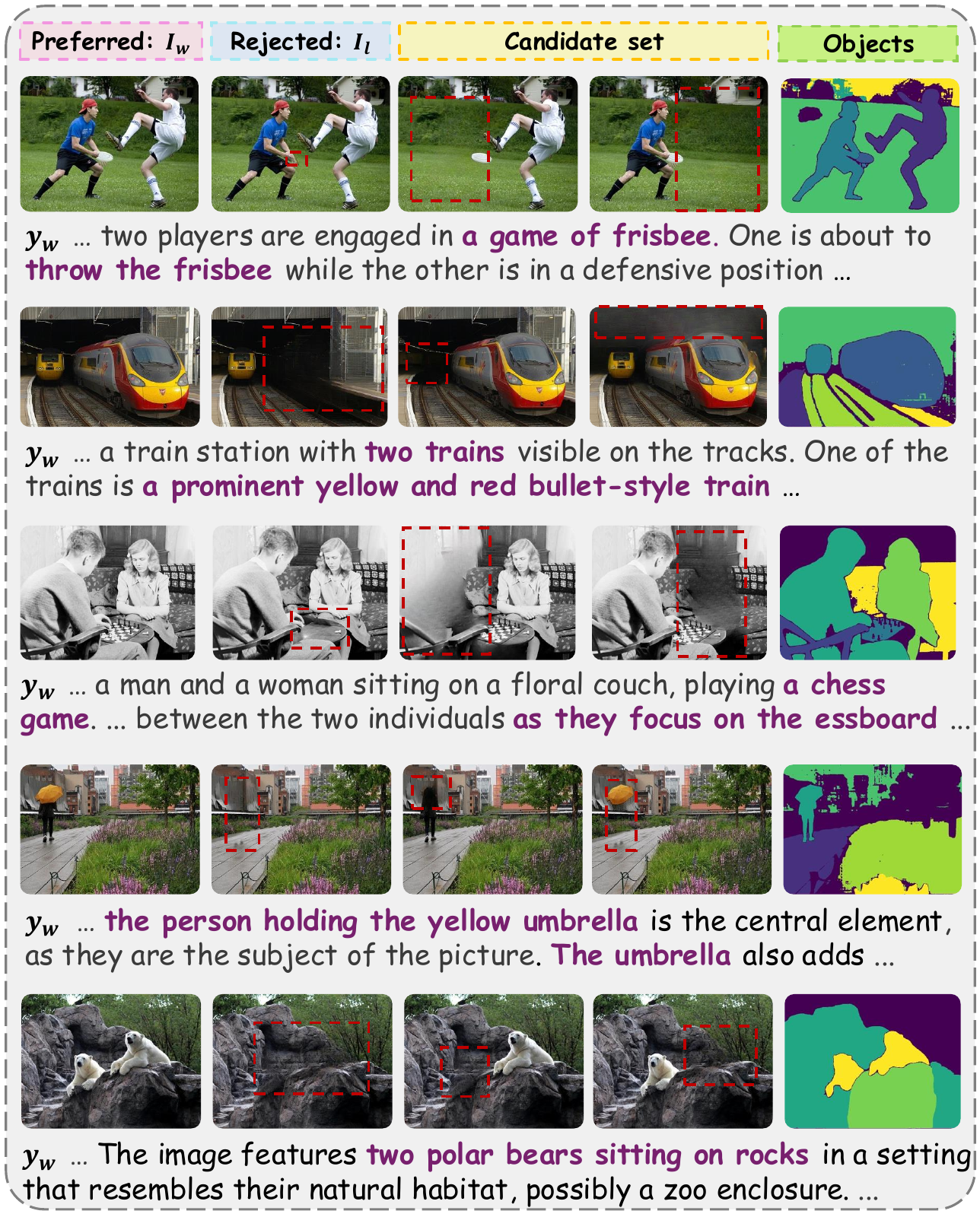}}
\caption{Visualization results of our vision-based rejected image construction, where we present the rejected image and the candidate set, together with the segmentation set of the image.}
\label{fig:visualization}
\end{center}
\vskip -0.3in
\end{figure}

\noindent \textbf{Influences of vision-based preference pair construction.}
To demonstrate the effectiveness of vision-based preference pairs construction, we present several instances of the rejected image $\mathcal{I}_{l}$, together with the candidate set, and the segmented objects in Figure~\ref{fig:visualization}. The visualization results reveal that: (1) by leveraging cascaded visual foundation models, we precisely localize and remove visual elements while preserving the underlying image structure, and (2) comparing with the preferred response $y_w$ and the original image $\mathcal{I}_{w}$, even subtle removals of elements can result in significant semantic differences from $y_w$, demonstrating our efficacy in identifying critical visual details.

\noindent \textbf{Effects of our adaptive optimization strategy.} 
We evaluate different optimization strategies by employing the discriminative tasks of the AMBER benchmark, reporting the F1 score of the fine-grained attribute ($\text{F1}_{\text{A}}$), relation ($\text{F1}_{\text{R}}$), and existence ($\text{F1}_{\text{E}}$) hallucination. Additionally, we also present the F1 score and accuracy across the entire 15K sample dataset.
The experimental results are illustrated in Table~\ref{tab:ab1-language-vision}, where ``Language DPO'' refers to Direct Preference Optimization (DPO) \cite{DPO} with language-based preferences, ``+Vision'' denotes DPO with equal weight for the vision- and language-based preferences, and ``+ Adaptive'' corresponds to our adaptive strategy.  Our experimental findings reveal that: (1) Simply combining the vision- and language-based preferences leads to performance degradation, evidenced by the significant drops in the F1 score and accuracy between ``DPO'' and ``+Vision'', especially for the accuracy, where it achieves 51.8\% accuracies, obtains more than 20\% performance drops with ``DPO''.
(2) Our adaptive strategy demonstrates substantial performance improvements across all dimensions, substantiating its effectiveness in dynamically aggregating vision- and language-based preferences.

\subsection{Performance Comparisons (RQ2)}
\begin{figure}[t]
\begin{center}
\centerline{\includegraphics[width=\columnwidth]{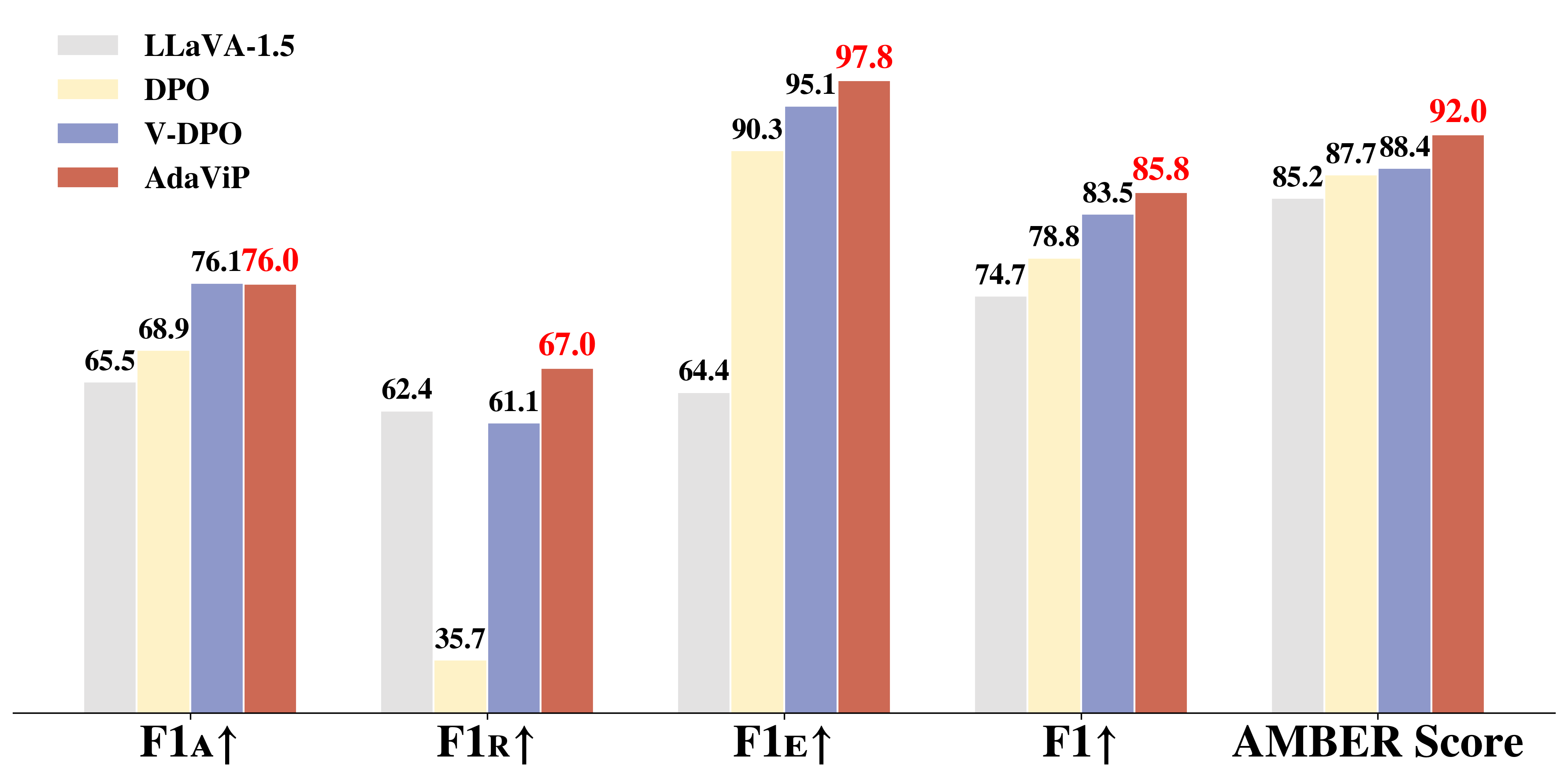}}
\caption{Performance comparisons of the fine-grained metrics in AMBER bench, where $\text{F1}_{\text{A}}$, $\text{F1}_{\text{R}}$, $\text{F1}_{\text{E}}$ represents the Attribute, Relation, and Existence hallucination on the discriminative benchmark, respectively. $\text{F1}$ measures the overall F1 scores on the discriminative benchmark, and $\text{AMBER Score}$ reflects the average performance on both generative and discriminative benchmarks.}
\label{fig:bar_plot}
\end{center}
\vskip -0.3in
\end{figure}

In this section, we conduct a comparative analysis of our proposed method against state-of-the-art baselines using three trustworthiness benchmarks: Object HalBench, AMBER, and MMHal-Bench. Our method is evaluated against a diverse set of baselines, including hallucination-specific, preference optimization-based, and the proprietary baseline GPT-4V. The evaluation is performed on both LLaVA-1.5 7B and 13B models. The experimental results, alongside those of the vanilla DPO, are summarized in Table \ref{tab:comparison_SOTA}.
We draw the following observations:
\newline
\noindent $\bullet$ Based on LLavA-1.5 7B and 13B, our AdaViP demonstrates \textit{consistent superiority} over DPO in mitigating hallucination across all three benchmarks. This underscores that AdaViP, which incorporates vision-based preference pairs and adaptive preference optimization, can effectively enhance preference alignment performance. It's noteworthy that, on AMBER Generative, compared with language-only DPO, it achieves a substantial improvement of over 70\% for both the 7B and 13B models, which is quite notable.
\newline
\noindent $\bullet$ AdaViP achieves new state-of-the-art performance in trustworthiness among various benchmarks, outperforming other open-source models by a large margin. Furthermore, based on LLaVA-1.5 7B and 13B, our AdaViP demonstrates a lower hallucination rate compared to GPT-4 on Object HalBench, AMBER Generative, and AMBER Score. For instance, on the Object HalBench, we decrease the hallucination rates by 4.6\% and 2.3\%  at the response and mention level, respectively, significantly surpassing GPT-4.
\newline
\noindent $\bullet$ We also conduct fine-grained evaluations on the AMBER Discriminative benchmark to verify the effectiveness of AdaViP over the vision-based method V-DPO \cite{V-dpo} in Figure \ref{fig:bar_plot}, where the F1 result corresponding to attribute ($\text{F1}_{\text{A}}$), relation ($\text{F1}_{\text{R}}$), existence ($\text{F1}_{\text{E}}$), and overall ($\text{F1}$) hallucination are illustrated. We also calculate the AMBER Score via $(100-CHAIR+F1)/2$. 
Compared with V-DPO, our AdaViP not only achieves comparable performances across the fine-grained tasks but also improves 2.3\% and 3.6\% on the overall F1 and AMBER Score, respectively.

\subsection{In-depth Analysis (RQ3)}
\begin{figure}[t]
\begin{center}
\centerline{\includegraphics[width=\columnwidth]{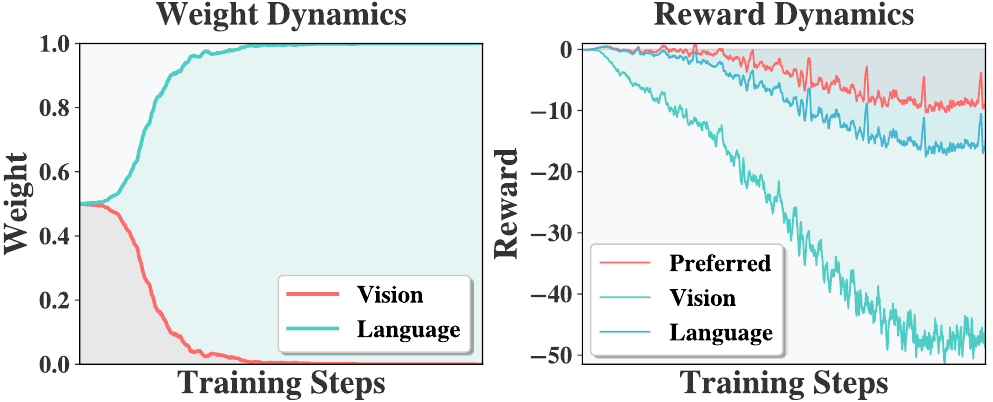}}
\caption{The dynamic weight between the vision- and language-based preferences and the corresponding reward dynamics for different samples during the training procedure.}
\label{fig:adaptive_weight}
\end{center}
\vskip -0.3in
\end{figure}

In this section, we analyze the weight and corresponding reward dynamics behind our adaptive preference optimization procedure. The dynamics during the training steps are shown in Figure~\ref{fig:adaptive_weight}. The observation can be summarized as:
\newline
\noindent $\bullet$ From the Weight Dynamics, the weights for vision-based preferences drop rapidly, while the weights for language-based ones increase and dominate the later training process. This suggests that the model progressively shifts its focus towards language-based preferences, and the model can effectively capture the visual context in the early stage.
\newline
\noindent $\bullet$ From the Reward Dynamics, the reward gap between the preferred and language-based rejected samples fluctuates during early training steps, but becomes stable and enlarged towards the end. This implies that the model can adaptively leverage both visual and language-based preferences, starting with a reliance on vision, and then transitioning towards an effective integration of both modalities.

\begin{table}[t]
\caption{Experimental results over the discriminative benchmark of AMBER for difference components of our AdaViP, where $\text{F1}_{\text{A}}$, $\text{F1}_{\text{R}}$, $\text{F1}_{\text{E}}$ represents the Attribute, Relation, and Existence hallucination. $\text{F1}$ and $\text{Acc.}$ measures the overall F1 score and Accuracy.}
\label{tab:ab1-language-vision}
\begin{center}
\begin{sc}
\scalebox{0.87}{
\begin{tabular}{l | c c c c c}
\toprule
\multirow{2}{*}{\textbf{Method}} & \multicolumn{5}{c}{\textbf{Discriminative}} \\
& \textbf{F1\textsubscript{E}$_{\uparrow}$} & \textbf{F1\textsubscript{A}$_{\uparrow}$} & \textbf{F1\textsubscript{R}$_{\uparrow}$} & \textbf{F1${\uparrow}$} & \textbf{Acc.$\uparrow$}\\
\midrule
LLaVA-1.5-7B & 64.4 & 65.5 & 62.4 & 74.7 & 73.9 \\
DPO & 90.3 & 68.9 & 35.7 & 78.8 & 64.8\\
\midrule
+ Vision & 85.1 & 61.2 & 47.4 & 72.6 & 51.8\\
\rowcolor{lightgray} + Adaptive & \textbf{97.8} & \textbf{76.0} & \textbf{67.0} & \textbf{85.8} & \textbf{79.9} \\
\bottomrule
\end{tabular}
}
\end{sc}
\end{center}
\vskip -0.4cm
\end{table}

\section{Related Works}
\label{sec: related_works}

In this section, we overview the related background. Specifically, we first briefly summarize recent hallucination mitigation methods for Multi-modal Large Language Models (MLLMs), and then we discuss strategies to align with human preferences. Finally, we enumerate the differences compared with related methods.

\subsection{Hallucination mitigation in MLLMs}
Hallucination, as a key indicator of trustworthiness, refers to the fact that the MLLM outputs are not aligned with the image content \cite{hallucination-survey}. Current works to mitigate hallucination can be summarized into the following aspects: 

\noindent \textbf{$\bullet$ Data Cleaning.} Recent works like \cite{LRV-Instruction, Hallucidoctor} assume that the noises in the curated SFT data might lead to hallucination, they propose data-cleaning strategies to address this. \newline
\noindent \textbf{$\bullet$ Enriched Visual Representation.} Works like \cite{Vcoder, MMVP} observe that the visual representation in the current MLLM is insufficient, which might result in hallucination, thus, they introduce strategies to enrich the visual representation. \newline
\noindent \textbf{$\bullet$ Test-time Augmentation.} Current works such as \cite{VCD, CFG} propose MLLM-based augmentation strategies to enhance the model's responsiveness to visual details. \newline
\noindent \textbf{$\bullet$ Preference Alignment.} By collecting hallucination-based preference data, works like \cite{RLHF-V, LLaVA-RLHF} transfer the alignment strategies from LLM into multi-modal scenarios, significantly mitigating the hallucination issue.

\subsection{Alignment with Human Preference}
\noindent \textbf{Alignment in Large Language Models (LLMs).} 
Current alignment strategies have demonstrated effectiveness in aligning the model outputs with human preferences. As a pioneering work, RLHF \cite{rlhf} proposes a two-stage strategy, which firstly trains a parametric reward model, and then optimizes the preference utilizing the reward with PPO \cite{PPO}. To simplify such a process and reduce computational complexity, subsequent works like \cite{DPO} and \cite{SimPO} replace the parametric reward with an implicit function, and further remove the reference model, respectively.

\noindent \textbf{Alignment in Multi-modal LLMs.}
To transfer the alignment paradigm into MLLMs, current methods explore from the following two aspects: 
(1) Preference Data Curation. Similar to LLMs, works like \cite{RLHF-V, RLAIF-V} construct preference data by employing larger models \cite{LLaVA-Next}, or human efforts.
(2) Visual detail enhancement. Recent works, such as \cite{mDPO, FiSAO}, identify that current methods lack sufficient attention on visual details. To remedy this, they propose strategies that either destroy the given image \cite{V-dpo, mDPO} or emphasize the tokens corresponding to key visual elements \cite{FiSAO, TPO}.

\noindent \textbf{Differences:} 
Our method belongs to the ``visual detail enhancement'', and offers the following advantages over similar methods \cite{mDPO, POVID, V-dpo}: 
Existing methods degrade the overall structure of the image by randomly cropping, adding noises, replacement, which inevitably introduces random noises into the preference pairs. Conversely, we preserve the integrity of the image by selectively removing only several key elements.
Moreover, the challenge to balance the vision- and language-based preference pairs remains not yet explored, in this paper, we introduce an adaptive balancing strategy and demonstrate significant effectiveness.
\section{Conclusion}
\label{sec:conclusion}

In this paper, we propose AdaViP, an \textbf{Ada}ptive \textbf{Vi}sion-enhanced \textbf{P}reference optimization approach, to improve multi-modal large language models' (MLLMs) sensitivity to the critical visual context. Our contributions are twofold: (1) a vision-based preference pair construction that enhances the model's focus on visual details by precisely extracting and removing key visual elements in the image; and (2) an adaptive preference optimization to dynamically balance the vision- and language-based preferences, ensuring a more robust multi-modal alignment. Significant performance gains across three distinct benchmarks demonstrate the effectiveness of our AdaViP.

\noindent \textbf{Limitations and Future Work.} In this paper, our focus is on images and text; future work will explore scalability to broader applications, such as video and audio, leveraging advanced foundation models across different modalities to comprehensively demonstrate our practical impact.

\normalem
{
    \small
    \bibliographystyle{ieeenat_fullname}
    \bibliography{main}
}

\end{document}